\documentclass[10pt,letterpaper]{article}

\usepackage[margin=1in]{geometry}
\usepackage{times}
\usepackage{graphicx}
\usepackage{booktabs}
\usepackage{amsmath}
\usepackage{amssymb}
\usepackage{hyperref}
\usepackage{xcolor}
\usepackage[font=small,labelfont=bf]{caption}
\usepackage{enumitem}
\usepackage{natbib}
\usepackage{microtype}

\title{Can Vision-Language Models See Squares?\\Text-Recognition Mediates Spatial Reasoning Across Three Model Families}

\author{Yuval Levental\\Rochester Institute of Technology\\
\texttt{yhl3051@g.rit.edu}}

\date{}

\begin{document}
\maketitle

% ============================================================
% ABSTRACT
% ============================================================
\begin{abstract}
We present a simple experiment that exposes a fundamental limitation in vision-language models (VLMs): the inability to accurately localize filled cells in binary grids when those cells lack textual identity. We generate fifteen $15\times15$ grids with varying density (10.7\%--41.8\% filled cells) and render each as two image types---text symbols (\texttt{.} and \texttt{\#}) and filled squares without gridlines---then ask three frontier VLMs (Claude Opus, ChatGPT 5.2, and Gemini 3 Thinking) to transcribe them. In the text-symbol condition, Claude and ChatGPT achieve $\sim$91\% cell accuracy and $\sim$84\% F1, while Gemini achieves 84\% accuracy and 63\% F1. In the filled-squares condition, all three models collapse to 60--73\% accuracy and 29--39\% F1. Critically, all conditions pass through the same visual encoder---the text symbols are images, not tokenized text. The text-vs-squares F1 gap ranges from 34 to 54 points across models. A symbol-type ablation reveals this gap is graded, not binary: Unicode square characters ($\square\blacksquare$) produce intermediate performance (69--77\% F1), and embedding text labels inside filled squares recovers performance for Claude and Gemini (80--86\% F1) but not ChatGPT (51\% F1), demonstrating model-specific text-visual pathway interactions. Each model exhibits a distinct failure mode in the squares condition---systematic under-counting (Claude), massive over-counting (ChatGPT), and template hallucination (Gemini)---but all share the same underlying deficit: severely degraded spatial localization for non-textual visual elements.
\end{abstract}

% ============================================================
% 1. INTRODUCTION
% ============================================================
\section{Introduction}

Vision-language models (VLMs) such as Claude, GPT-4o, and Gemini have demonstrated impressive capabilities across visual understanding tasks---describing images, interpreting charts, reasoning about diagrams, and answering questions about visual content. These successes have led to an implicit assumption that VLMs possess robust spatial reasoning over arbitrary visual inputs.

We challenge this assumption with a deliberately simple experiment. We construct fifteen $15\times15$ binary grids---each cell either filled (black) or empty (white)---and ask VLMs to transcribe them. The grids span fill densities from 10.7\% to 41.8\%, and each grid is rendered in two visual formats: as text characters (\texttt{.} and \texttt{\#}) in a monospace font, and as filled squares without gridlines. Both are presented as PNG images; neither bypasses the visual encoder.

We evaluate three frontier VLMs from three different organizations: Claude Opus (Anthropic), ChatGPT 5.2 (OpenAI), and Gemini 3 Thinking (Google). The results are stark and consistent. When grids are encoded as text symbols, Claude and ChatGPT achieve approximately 91\% cell accuracy and 84\% F1, while Gemini achieves 84\% accuracy. When the same information is encoded as filled squares, all three models collapse: accuracy drops to 60--73\% and F1 to 29--39\%. The text-vs-squares F1 gap ranges from 34 to 54 points across models.

This gap is surprising because the information content is identical across conditions. The only difference is whether the visual encoding maps to recognizable text characters. We hypothesize that VLMs behave as if they rely on a text-recognition pathway---essentially performing internal OCR---to achieve high-fidelity spatial reasoning, and that their native visual spatial reasoning over non-textual elements is far weaker than commonly assumed. The replication of this gap across three independent model families, each with different architectures, training data, and visual encoders, suggests this is a fundamental property of current VLM designs rather than an idiosyncrasy of any single system.

% ============================================================
% 2. RELATED WORK
% ============================================================
\section{Related Work}

\paragraph{Spatial Reasoning Benchmarks.}
A growing body of work evaluates VLM spatial reasoning. \citet{liu2023vsr} introduced VSR (Visual Spatial Reasoning), testing spatial relationships between objects in natural images. SpatialBench \citep{cai2024spatialbench} evaluates 3D spatial reasoning including metric depth estimation, proximity, and counting. \citet{chen2024spatialvlm} proposed SpatialVLM, demonstrating that VLMs' limited spatial reasoning stems from insufficient 3D spatial data in training rather than architectural limitations. \citet{cheng2024spatialrgpt} introduced SpatialRGPT, which enhances spatial perception through depth-augmented region-level reasoning. More recently, \citet{liu2025deconstructing} provided a comprehensive survey evaluating 37 models across 9 benchmarks, revealing persistent weaknesses in fine-grained spatial configurations. Spatial-DISE (2025) benchmarked 28 models under a cognitive framework and found universal performance ceilings. TransformEval (2025) specifically probed spatial transformation reasoning and found catastrophic failures in compositional spatial tasks. iVISPAR \citep{ramakrishnan2025ivispar} introduced an interactive sliding-puzzle benchmark and found that VLMs consistently fall short of human performance.

Our work complements these efforts but differs in a key respect: rather than testing spatial reasoning over complex natural or 3D scenes, we use the simplest possible visual stimulus---a binary grid---to isolate spatial localization from semantic understanding.

\paragraph{Counting and Fine-Grained Localization.}
VLMs are known to struggle with counting tasks. \citet{radford2021clip} noted that CLIP struggles with counting objects and fine-grained classification. \citet{paiss2023teaching} specifically addressed teaching CLIP to count. \citet{zhang2024clip} studied quantity bias in CLIP. \citet{hu2025exposing} systematically exposed VLM failures in compositional counting. Our binary grid task can be viewed as the simplest possible counting and localization task---yet VLMs still fail dramatically when cells are rendered as filled squares.

\paragraph{Visual Encoder Limitations.}
CLIP \citep{radford2021clip}, the visual encoder underlying many VLMs, was trained via image-text contrastive learning that optimizes for global semantic alignment rather than fine-grained spatial features. \citet{tong2024eyes} identified CLIP-blind spots---perceptually distinct image pairs that CLIP confuses. \citet{wei2024clip} showed that CLIP's deep-layer features focus on global properties while neglecting local pixel-level detail. \citet{zhou2025cloc} proposed CLOC to add region-text contrastive objectives. \citet{jing2025clipin} introduced CLIP-IN to enhance fine-grained visual understanding. These efforts confirm that CLIP's image-level training produces encoders with limited spatial precision.

\paragraph{OCR and Document Understanding.}
The relationship between OCR capabilities and VLMs is directly relevant to our findings. Modern VLMs perform implicit OCR when processing text in images. DocVLM \citep{nacson2024docvlm} explicitly integrates an OCR encoder into VLMs, demonstrating that text-recognition and visual-understanding pathways serve complementary roles. Our results suggest that text in images activates a high-fidelity recognition pathway that preserves spatial position, while non-textual visual elements rely on a lower-fidelity pathway.

% ============================================================
% 3. EXPERIMENTAL SETUP
% ============================================================
\section{Experimental Setup}

\subsection{Grid Generation}

We generate 15 binary grids of size $15\times15$ (225 cells each). Grids are ordered by density, with filled-cell counts ranging from 24 (Grid 1, 10.7\%) to 94 (Grid 15, 41.8\%). Filled cells are distributed to include both isolated cells and clusters of adjacent cells, providing a range of localization difficulty.

\subsection{Visual Encoding Conditions}

Each grid is rendered as a PNG image under two conditions:

\begin{description}[leftmargin=1.5em,labelindent=0em]
\item[Text symbols:] Each cell is rendered as \texttt{.} (empty) or \texttt{\#} (filled) using a monospace font (DejaVu Sans Mono, 28pt). The resulting image is a photograph of text, processed by the visual encoder like any other image.
\item[Unsegmented squares:] Each cell is rendered as a filled (black) or empty (white) square with no gridlines. Adjacent filled cells merge into contiguous black regions.
\end{description}

We also ran a segmented-squares condition (with gridlines) on Claude Opus; this yielded results nearly identical to unsegmented squares (32.6\% vs 29.6\% F1), indicating gridlines provide negligible benefit. For parsimony, the cross-model evaluation uses only text and unsegmented conditions.

\subsection{Models and Prompting}

We evaluate three frontier VLMs from three organizations:

\begin{description}[leftmargin=1.5em,labelindent=0em]
\item[Claude Opus] (claude-opus-4-6, Anthropic) via claude.ai web interface.
\item[ChatGPT 5.2] (OpenAI) via chatgpt.com Plus Auto mode.
\item[Gemini 3 Thinking] (Google) via gemini.google.com with extended thinking enabled.
\end{description}

Each model is given a fixed instruction prompt specifying the task format, then presented with grids in batches of five. The prompt explicitly prohibits the use of code or image-processing tools, requiring transcription from visual inspection alone. Each condition is run in a separate, clean session.

\subsection{Metrics}

We report cell accuracy (fraction of 225 cells correctly classified) and black-cell F1 (harmonic mean of precision and recall for filled-cell detection). Aggregate metrics are computed over all 3,375 cells ($15 \text{ grids} \times 225 \text{ cells}$). F1 is the more informative metric because accuracy is inflated by the large number of easily-classified empty cells.

% ============================================================
% 4. RESULTS
% ============================================================
\section{Results}

\subsection{Single-Model Results (Claude Opus)}

Table~\ref{tab:claude} shows Claude Opus results across all conditions tested. The text-symbol condition dramatically outperforms both square conditions on every metric. The two square conditions perform similarly, indicating that gridlines provide negligible benefit for spatial localization.

\begin{table}[h]
\centering
\caption{Claude Opus results across encoding conditions.}
\label{tab:claude}
\begin{tabular}{lcccc}
\toprule
Condition & Cell Acc. & Precision & Recall & F1 \\
\midrule
Text symbols & 90.6\% & 83.3\% & 83.4\% & 83.4\% \\
Segmented squares & 64.9\% & 33.6\% & 31.7\% & 32.6\% \\
Unsegmented squares & 66.9\% & 34.5\% & 25.9\% & 29.6\% \\
\bottomrule
\end{tabular}
\end{table}

\subsection{Cross-Model Replication}

To test whether the text-vs-squares gap is specific to Claude or reflects a general VLM limitation, we replicate the experiment on ChatGPT 5.2 and Gemini 3 Thinking. Table~\ref{tab:crossmodel} summarizes results across all three models.

\begin{table}[h]
\centering
\caption{Cross-model comparison. The text-vs-squares F1 gap replicates across all three model families.}
\label{tab:crossmodel}
\begin{tabular}{llcccc}
\toprule
Model & Condition & Cell Acc. & Prec. & Recall & F1 \\
\midrule
Claude Opus & Text & 90.6\% & 83.3\% & 83.4\% & 83.4\% \\
 & Unseg. Squares & 66.9\% & 34.5\% & 25.9\% & 29.6\% \\
 & \textbf{Gap} & & & & \textbf{$-$53.8} \\
\midrule
ChatGPT 5.2 & Text & 91.3\% & 85.8\% & 82.3\% & 84.0\% \\
 & Unseg. Squares & 65.9\% & 37.1\% & 41.5\% & 39.2\% \\
 & \textbf{Gap} & & & & \textbf{$-$44.8} \\
\midrule
Gemini 3 Think & Text & 83.9\% & 63.1\% & 62.5\% & 62.8\% \\
 & Unseg. Squares & 73.3\% & 50.8\% & 20.0\% & 28.7\% \\
 & \textbf{Gap} & & & & \textbf{$-$34.1} \\
\bottomrule
\end{tabular}
\end{table}

The core finding replicates robustly: all three models exhibit a dramatic F1 gap between text and squares conditions, ranging from 34.1 points (Gemini) to 53.8 points (Claude). Gemini's smaller gap reflects not superior square performance but rather degraded text performance---its text F1 (62.8\%) is 20+ points below Claude and ChatGPT, due to a density-dependent collapse discussed below.

\subsection{Density Effects Across Models}

\begin{figure}[ht]
    \centering
    \includegraphics[width=\textwidth]{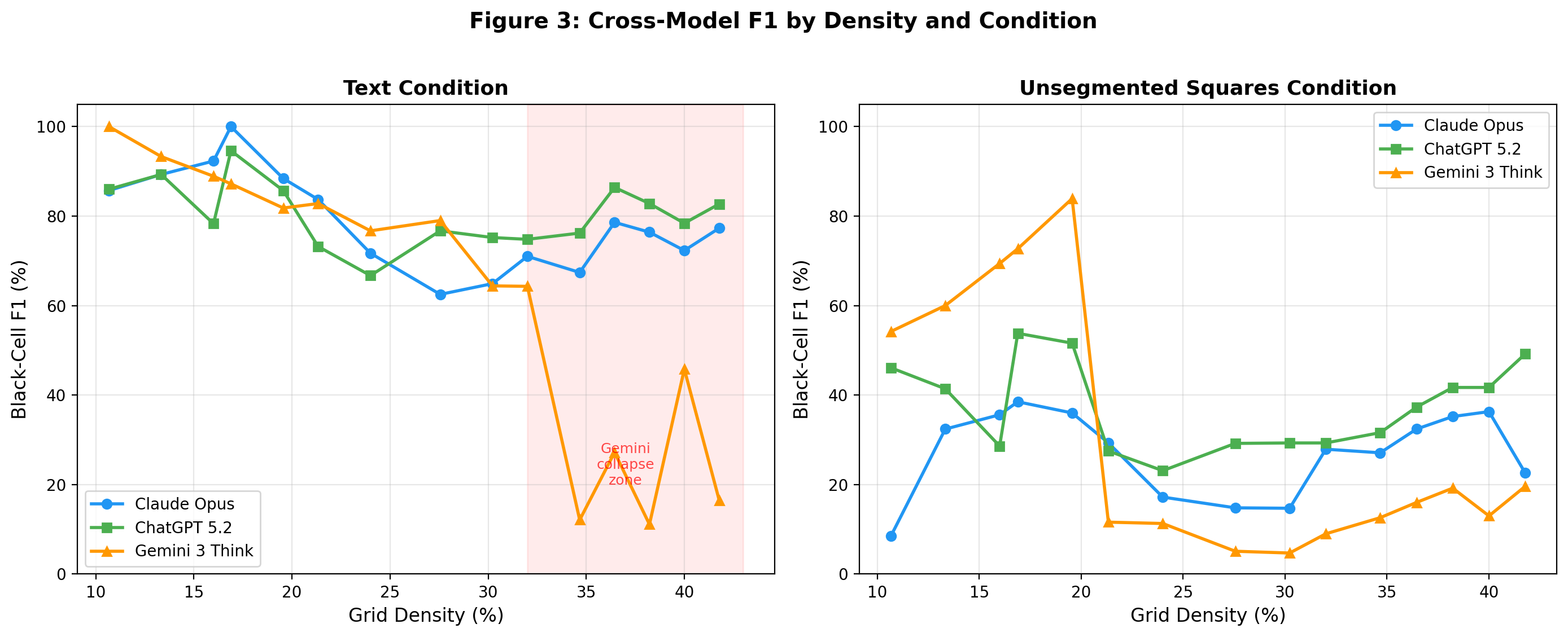}
    \caption{Black-cell F1 vs.\ grid density across models. Left: text condition. Right: unsegmented squares. Claude and ChatGPT maintain stable text performance across all densities; Gemini collapses above $\sim$32\%. All models show poor and unstable performance in the squares condition.}
    \label{fig:crossmodel_f1}
\end{figure}

Figure~\ref{fig:crossmodel_f1} reveals important differences in how each model handles increasing grid density. In the text condition (left panel), Claude and ChatGPT maintain stable F1 scores across the full density range (62--100\% and 66--95\%, respectively), with no systematic degradation. Gemini performs comparably at low densities (grids 1--10, $\leq$32\%: 64--100\% F1) but collapses catastrophically on dense grids (grids 11--15, $\geq$35\%: 11--46\% F1). At high density, Gemini generates hallucinated geometric patterns---simple cross or plus shapes---bearing no resemblance to the input, suggesting a capacity limit in its text-recognition pathway that causes it to fall back on pattern confabulation.

In the unsegmented squares condition (right panel), all three models show poor F1 overall, but a surprising pattern emerges at low density. Gemini achieves the highest squares F1 on every grid from 1--5 ($\leq$20\% density), averaging 68.0\% F1 versus 44.3\% for ChatGPT and 30.2\% for Claude. This suggests that Gemini's visual pathway is actually the strongest of the three at perceiving discrete, isolated visual objects---precisely the task that a visual encoder should excel at. However, Gemini's advantage reverses sharply above 20\% density: grids 6--15 collapse to 5--20\% F1, with the model generating identical L-shaped tetromino patterns tiled across all dense grids regardless of the actual input. By contrast, Claude and ChatGPT maintain more stable (if uniformly poor) squares performance across all densities. The pattern suggests a ``forest vs.\ trees'' distinction: Gemini is better at perceiving distinct filled regions as discrete visual objects, but has a sharper capacity ceiling, while Claude and ChatGPT compensate for weaker visual perception with more robust text-recognition pathways.

\begin{figure}[ht]
    \centering
    \includegraphics[width=0.7\textwidth]{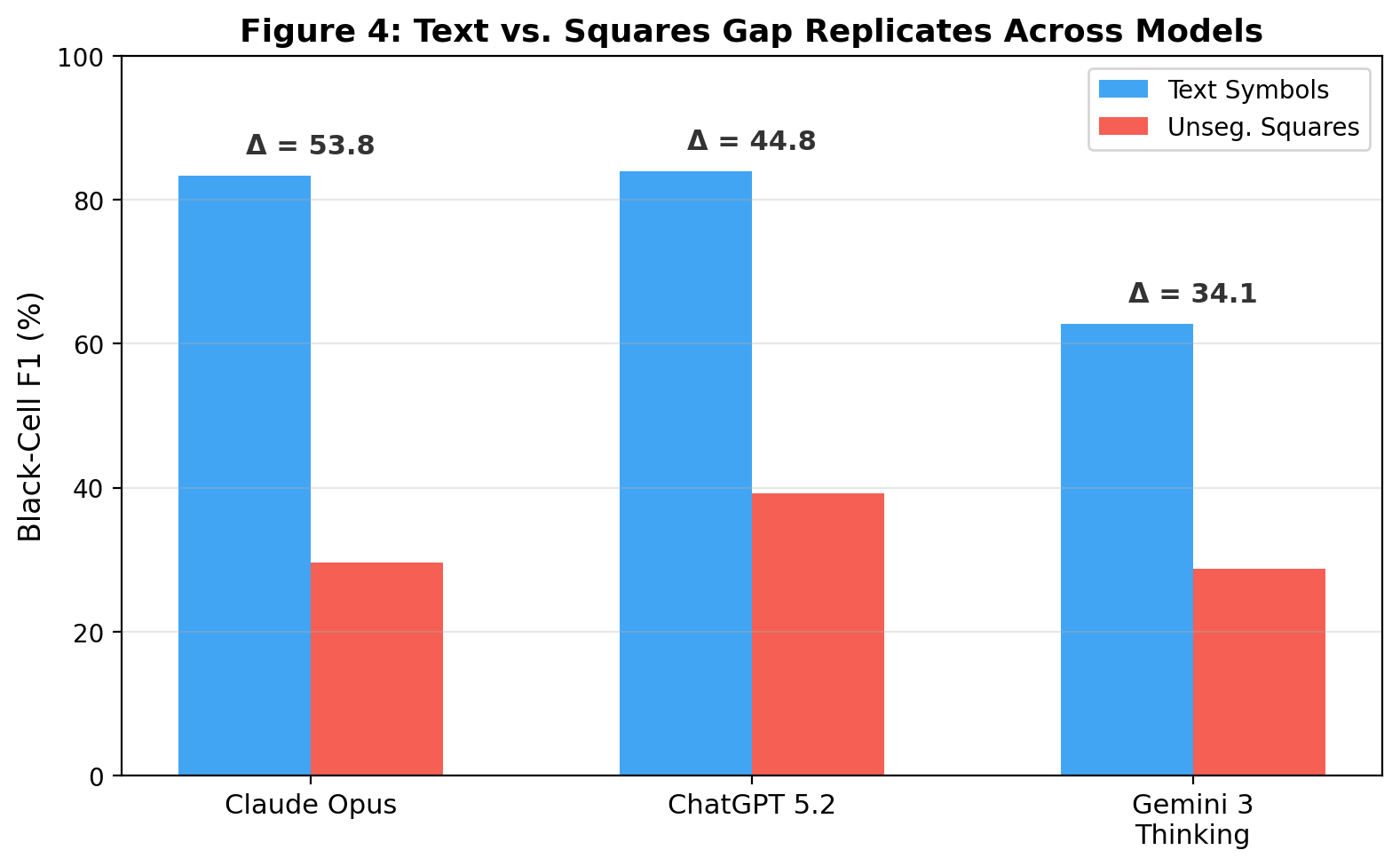}
    \caption{Text-vs-squares F1 gap replicates across all three model families. The gap ranges from 34 to 54 F1 points.}
    \label{fig:gap_bars}
\end{figure}

\subsection{Qualitative Failure Mode Analysis}

\begin{figure}[ht]
    \centering
    \includegraphics[width=\textwidth]{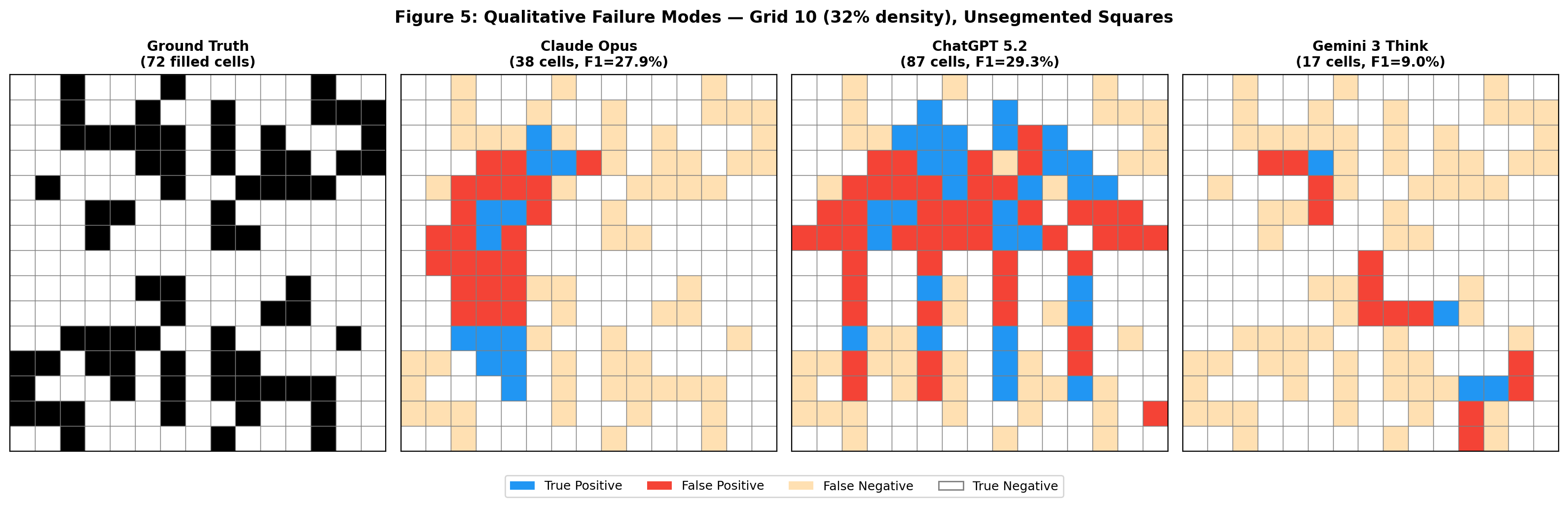}
    \caption{Qualitative failure modes on Grid 10 (32\% density, unsegmented squares). Blue: true positives. Red: false positives. Orange: false negatives. Each model fails differently but all exhibit severe spatial mislocalization.}
    \label{fig:failure_modes}
\end{figure}

Each model exhibits a qualitatively distinct failure mode in the squares condition, illustrated in Figure~\ref{fig:failure_modes}:

\paragraph{Claude Opus: Systematic under-counting.} Claude consistently transcribes fewer filled cells than exist (avg.\ 45 predicted vs.\ 60 actual). Even when approximate counts are correct, cells are displaced to incorrect positions. The model appears to detect the approximate \textit{region} of filled cells but cannot precisely localize individual cell boundaries.

\paragraph{ChatGPT 5.2: Massive over-counting.} ChatGPT dramatically over-predicts filled cells at high densities, transcribing 112--133 filled cells for grids with 78--94 true cells. The model also frequently loses grid dimensions, producing rows of 16--17 characters. It appears to hallucinate filled cells in the vicinity of true clusters, creating expanded, blurred versions of the actual pattern.

\paragraph{Gemini 3 Thinking: Template hallucination.} At moderate-to-high densities, Gemini abandons the input entirely and generates stereotyped geometric patterns---L-shapes, crosses, and plus signs---that bear no resemblance to the actual grid. In the text condition, high-density grids are transcribed as simple ASCII art (letters E, H, S). This suggests the model falls back on learned visual prototypes when the input exceeds its processing capacity.

The diversity of failure modes is itself informative: three independent architectures, trained on different data with different visual encoders, all fail at the same task but in different ways. This suggests the underlying limitation is not an idiosyncrasy of any specific training procedure but a structural property of the text-recognition-mediated approach to spatial reasoning.

\subsection{Ablation: Symbol Type Mediates the Gap}

Our main experiment establishes a large F1 gap between text-symbol and filled-square grids. A natural question is whether this gap reflects a binary distinction between text and non-text, or a more graded continuum. To investigate, we introduce two intermediate rendering conditions and test all three models on three grids spanning the density range (Grid~1 at 10.7\%, Grid~7 at 24.0\%, Grid~14 at 40.0\%).

\paragraph{Unicode squares (\texorpdfstring{$\square\blacksquare$}{white/black square}).} We render each cell using Unicode white square (U+2B1C) or black square (U+2B1B) characters. These are valid text tokens with Unicode codepoints, yet they visually resemble the filled-square condition. This tests whether text identity alone is sufficient for high-fidelity spatial reasoning.

\paragraph{Text-in-squares.} We render grids identically to the filled-square baseline, but place a ``1'' in white text inside each black cell and a ``0'' in light gray inside each white cell. This preserves the visual appearance of filled squares while reintroducing textual content that can anchor spatial reasoning.

Table~\ref{tab:ablation} reports F1 scores across all four conditions and all three models.

\begin{figure}[ht]
    \centering
    \includegraphics[width=\textwidth]{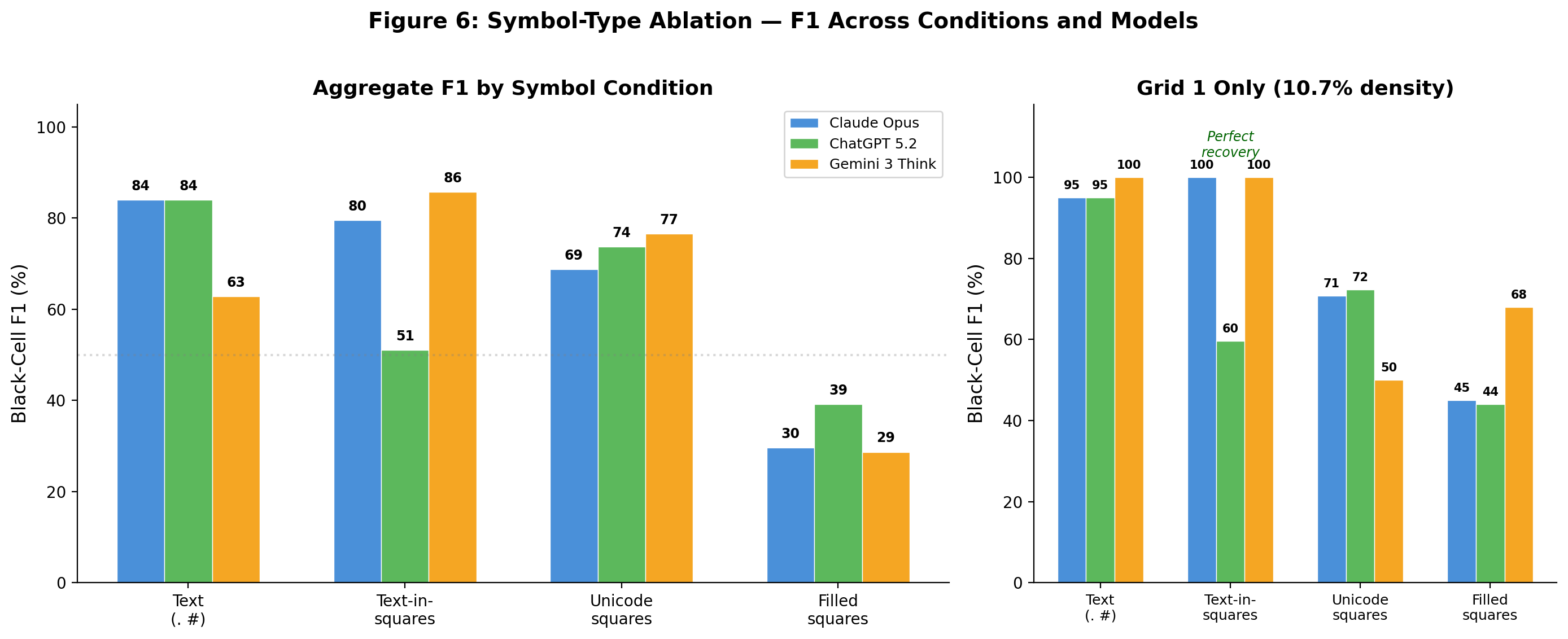}
    \caption{Symbol-type ablation. Left: aggregate F1 across three grids. The gap is graded, not binary. ChatGPT's text-in-squares degradation (51\%) reveals destructive pathway interference. Right: on sparse Grid~1, text-in-squares achieves 100\% F1 for Claude and Gemini---complete recovery of spatial reasoning.}
    \label{fig:ablation}
\end{figure}

\begin{table}[h]
\centering
\caption{Symbol-type ablation (F1, \%). Three grids tested per condition: Grid~1 (10.7\%), Grid~7 (24.0\%), Grid~14 (40.0\%). Text and filled-square baselines are from the corresponding grids in the full 15-grid evaluation.}
\label{tab:ablation}
\begin{tabular}{llcccc}
\toprule
Model & Condition & G1 & G7 & G14 & Agg. \\
\midrule
Claude Opus & Text (\texttt{.\#}) & $\sim$95 & $\sim$85 & $\sim$75 & $\sim$84 \\
 & Text-in-squares & \textbf{100.0} & 65.4 & 82.4 & 79.6 \\
 & Unicode squares & 70.8 & 75.0 & 64.3 & 68.8 \\
 & Filled squares & $\sim$45 & $\sim$30 & $\sim$20 & $\sim$35 \\
\midrule
Gemini 3 Think & Text (\texttt{.\#}) & $\sim$100 & $\sim$85 & $\sim$63 & $\sim$63 \\
 & Text-in-squares & \textbf{100.0} & 88.3 & 80.2 & 85.7 \\
 & Unicode squares & 50.0 & 85.2 & 78.5 & 76.6 \\
 & Filled squares & $\sim$68 & $\sim$15 & $\sim$10 & $\sim$29 \\
\midrule
ChatGPT 5.2 & Text (\texttt{.\#}) & $\sim$95 & $\sim$85 & $\sim$75 & $\sim$84 \\
 & Text-in-squares & 59.6 & 51.5 & 48.4 & 51.1 \\
 & Unicode squares & 72.3 & 69.2 & 76.7 & 73.7 \\
 & Filled squares & $\sim$44 & $\sim$30 & $\sim$30 & $\sim$39 \\
\bottomrule
\end{tabular}
\end{table}

The results reveal both a graded continuum and a striking model-specific divergence.

\paragraph{Text-in-squares recovers sparse-grid performance for Claude and Gemini.} Both models achieve 100\% F1 on Grid~1 when ``0''/``1'' labels are placed inside filled-square cells---complete recovery of spatial reasoning on sparse grids. For Gemini, text-in-squares produces the highest aggregate F1 of any condition (85.7\%), even exceeding its own text baseline (62.8\% over 15 grids). This demonstrates that the visual pathway's failure on filled squares is not due to inherent ambiguity in the stimulus; when text is present, even embedded within large black regions, these models can localize cells precisely.

\paragraph{ChatGPT exhibits the opposite pattern.} Adding text labels inside squares \textit{degrades} ChatGPT's performance (51.1\% F1) below even the filled-square baseline (39.2\% over 15 grids, but higher on these specific grids). ChatGPT performs better with Unicode squares (73.7\%) than with text-in-squares, suggesting that the visual complexity of overlaid text on filled backgrounds actively interferes with ChatGPT's processing rather than aiding it. This may reflect differences in how ChatGPT's visual encoder handles text-on-background contrast versus isolated symbols.

\paragraph{Unicode squares produce intermediate performance across all models.} All three models achieve 69--77\% aggregate F1 on Unicode squares---substantially better than filled squares ($\sim$29--39\%) but worse than pure text ($\sim$63--84\%). Since $\square$ and $\blacksquare$ are valid text tokens, this degradation may reflect token frequency rather than visual interference: \texttt{.} and \texttt{\#} appear orders of magnitude more often in training data than Unicode square characters. An alternative explanation is that the visual similarity between Unicode squares and actual filled squares partially disrupts the text-recognition pathway; the two hypotheses are not mutually exclusive.

\paragraph{Behavioral note.} When presented with the text-in-squares condition, Gemini~3 Thinking first attempted to write Python code to threshold pixel values---producing a grid of all dots---before reverting to visual transcription. The text-in-squares stimulus apparently triggered a ``programmatic task'' heuristic, suggesting the model interpreted the embedded text as an invitation to use tools rather than visual perception. This is itself evidence that text content in images activates qualitatively different processing strategies.

% ============================================================
% 5. DISCUSSION
% ============================================================
\section{Discussion}

\subsection{The Text-Recognition Pathway Hypothesis}

Our central finding is that VLMs can accurately localize visual elements when they are recognizable characters, but not when they are abstract filled regions---even though both are processed as pixel images through the same visual encoder. This gap replicates across three model families from three organizations, with F1 differences of 34--54 points.

We propose that VLMs behave \textit{as if} they possess two implicit processing pathways for spatial information. (We emphasize ``as if'' because we cannot directly observe internal model mechanisms from behavioral data alone; the following is a functional description, not an architectural claim.)

\begin{description}[leftmargin=1.5em,labelindent=0em]
\item[A text-recognition pathway] that identifies characters in images, maps them to discrete tokens, and preserves their sequential (and therefore spatial) positions with high fidelity. This pathway essentially performs internal OCR, converting visual information into the linguistic domain where the model has strong positional reasoning.
\item[A visual-feature pathway] that encodes non-textual visual content into latent representations optimized for semantic understanding. This pathway captures approximate spatial relationships (e.g., ``cluster of dark pixels in the upper-right region'') but loses precise coordinate-level information.
\end{description}

Our symbol-type ablation (Section~4.5) refines this model. Rather than a binary switch, performance depends on a graded ``text-recognition confidence'' that reflects at least three factors: (1)~whether the stimulus contains recognizable text at all (the dominant factor, accounting for $\sim$35--50 F1 points), (2)~the frequency of specific text tokens in training data (Unicode squares degrade $\sim$10--15 points relative to ASCII, possibly due to lower training-data frequency), and (3)~model-specific interactions between text and visual pathways (ChatGPT's degradation on text-in-squares suggests its pathways can interfere destructively when both are activated).

Importantly, the text-recognition pathway is not uniformly robust across models. Claude and ChatGPT maintain stable text-condition performance across all tested densities, while Gemini's text pathway collapses at moderate density ($\geq$32\%), falling back on pattern confabulation. This suggests that text-recognition pathway capacity varies across architectures and may depend on specific training decisions regarding OCR data, visual encoder design, or tokenization strategy.

Conversely, the visual-feature pathway is not uniformly weak. Gemini achieves the highest squares-condition F1 of any model on sparse grids ($\leq$20\% density), averaging 68\% F1 versus 44\% for ChatGPT and 30\% for Claude. This suggests that Gemini's visual encoder is actually the most spatially precise of the three when processing discrete, isolated visual objects. However, this advantage disappears above 20\% density, where Gemini collapses more severely than either competitor. The overall picture is that the three models occupy different positions in a tradeoff between text-pathway robustness and visual-pathway capacity: Claude and ChatGPT have strong text pathways but weaker visual pathways, while Gemini has a stronger visual pathway but a more brittle text pathway. In all cases, the text pathway substantially outperforms the visual pathway when it functions correctly.

\subsection{Alternative Explanations}

We consider and argue against two alternative explanations for our findings.

\paragraph{Image downsampling.} VLM visual encoders typically resize input images to a fixed resolution (e.g., $224\times224$ or $336\times336$ pixels, sometimes with tiling schemes for higher resolution). One might hypothesize that downsampling renders individual cells or characters unresolvable, particularly at high densities. However, this explanation fails on several counts. First, the text images have identical pixel dimensions regardless of density---a $15\times15$ grid of monospace characters occupies the same image size whether 10\% or 42\% of cells are filled. If downsampling blurred characters to illegibility, it would affect all grids equally, yet Claude and ChatGPT maintain $\geq$83\% accuracy across all densities. Second, downsampling applies equally to both conditions and therefore cannot explain the text-vs-squares gap within any single model. Third, Gemini's density-dependent collapse is \textit{bimodal}---performance drops from $\sim$100\% to $\sim$11\% F1 across a narrow density range---rather than exhibiting the smooth degradation that resolution loss would predict. Finally, the hallucinated outputs (ASCII art letters, repeating geometric templates) are structured and confident, not the noisy or incomplete responses expected from perceptual ambiguity.

\paragraph{Sequence length or decoding constraints.} Dense grids require long, highly constrained output sequences. Under uncertainty about the visual input, a language model may prioritize global regularity over local fidelity---generating patterns that are internally coherent but unrelated to the stimulus. This explanation likely contributes to Gemini's template hallucination specifically, but it cannot account for the text-vs-squares gap: both conditions require identical output lengths (15 rows of 15 characters), yet only the squares condition triggers the failure. The critical variable remains whether the visual input is recognizable as text, not the difficulty of generating the output sequence.

\subsection{Implications}

This finding has practical implications for applications relying on VLM spatial reasoning over non-textual visual elements. Medical imaging, autonomous systems, document analysis beyond text, and scientific visualization all require the visual-feature pathway that our results show is spatially imprecise across all tested models. Performance on text-heavy benchmarks may substantially overestimate spatial reasoning capability for non-textual content.

\subsection{Toward Bridging the Gap}

Our ablation results directly test two bridging strategies proposed in Section~5.2, with mixed results:

\paragraph{Textual scaffolding works---for some models.} The text-in-squares condition embeds ``0''/``1'' labels inside the filled-square rendering, providing textual anchors without changing the visual layout. For Claude and Gemini, this substantially recovers spatial reasoning (79.6\% and 85.7\% F1, respectively), with perfect recovery on sparse grids. However, for ChatGPT, textual scaffolding \textit{backfires} (51.1\% F1), demonstrating that this strategy is not universally effective and that text-visual interactions are model-dependent. Practical applications of textual scaffolding should therefore be validated per model.

\paragraph{Token familiarity matters.} The Unicode squares condition shows that even valid text tokens degrade when they are low-frequency in training data ($\sim$69--77\% F1 vs.\ $\sim$84\% for ASCII). This suggests that textual scaffolding is most effective when using high-frequency tokens that the model's text-recognition pathway can confidently decode.

\paragraph{Output format variation.} Requiring coordinate-list outputs (e.g., ``filled: (2,4), (3,5), ...'') instead of grid transcriptions would test whether the failure lies in perception or in generating long structured sequences, helping disentangle encoder limitations from decoder limitations. This remains untested.

\paragraph{Architectural interventions.} Training visual encoders with explicit spatial-coordinate prediction objectives---rather than the image-caption matching objectives used by CLIP and similar models---could improve native visual spatial resolution. Additionally, our ablation results suggest that introducing discrete visual tokens (analogous to VQ-VAE/VQ-GAN visual codebooks) into VLM pipelines could bridge the gap by giving the text-recognition pathway access to non-textual visual content through learned discrete representations.

\subsection{Limitations and Future Work}

We evaluate three models in a single trial per condition. While our 15 grids provide 15 independent measurements per condition and the cross-model replication strengthens generalizability, multi-trial repetition would further strengthen statistical confidence. The symbol-type ablation (Section~4.5) uses three grids rather than the full fifteen; extending it to all grids and additional symbol types (e.g., drawn circles, stars) would provide a finer-grained map of the text-to-visual transition. Gemini 3 Thinking showed high variance across runs: a second trial produced earlier collapse (at 21\% density vs.\ 35\%) with worse overall performance. We report the best Gemini run; even this shows the fundamental text-vs-squares gap. The run-to-run variance is itself informative, suggesting that Gemini's text-recognition pathway operates near a capacity boundary where small perturbations cause large performance swings.

All models were evaluated via web interfaces, which may apply hidden preprocessing (compression, resizing) to uploaded images. API-based evaluation with controlled image parameters would eliminate this potential confound. We note, however, that any such preprocessing applies equally to both conditions, so it cannot explain the text-vs-squares gap.

We do not perform prompt sensitivity analysis; results may vary with different prompt formulations, though the consistency across three different models and prompting styles suggests robustness. Evaluation of open-source VLMs with known architectures (e.g., LLaVA, Qwen-VL) would allow correlating behavioral findings with specific visual encoder properties, and would improve reproducibility. Testing across grid sizes (e.g., $10\times10$ to $30\times30$) would reveal whether the text-pathway advantage has a capacity ceiling. These extensions represent natural future work.

% ============================================================
% 6. CONCLUSION
% ============================================================
\section{Conclusion}

We demonstrate that three frontier VLMs from three organizations---Claude Opus, ChatGPT 5.2, and Gemini 3 Thinking---all exhibit a dramatic performance gap between text-symbol and filled-square encodings of identical binary grids, despite both being processed as images through the same visual encoder. The F1 gap ranges from 34 to 54 points across models. A symbol-type ablation reveals this gap is graded: Unicode square characters produce intermediate performance (69--77\% F1), and embedding text labels inside filled squares recovers spatial reasoning for two of three models---achieving 100\% F1 on sparse grids---but paradoxically degrades performance for ChatGPT, revealing model-specific text-visual pathway interactions.

Each model fails differently in the squares condition (under-counting, over-counting, template hallucination), but all share the same underlying deficit: severely degraded spatial localization for non-textual visual elements. Our experiment is deliberately simple---fifteen $15\times15$ binary grids, four rendering conditions, three models---yet the results are unambiguous and replicate across independent model families. Current VLMs behave as if much of their spatial reasoning is mediated by text recognition; when visual elements have no textual analog, localization degrades severely. We hope this finding motivates investigation into bridging the gap between text-mediated and native visual spatial reasoning in multimodal models.

% ============================================================
% REFERENCES
% ============================================================

\end{document}